\documentclass[letterpaper]{article} 
\usepackage{aaai25}  
\usepackage{times}  
\usepackage{helvet}  
\usepackage{courier}  
\usepackage[hyphens]{url}  
\usepackage{graphicx} 
\usepackage{natbib}  
\usepackage{caption} 
\usepackage{algorithm}
\usepackage{algorithmic}
\usepackage{booktabs}
\usepackage{xcolor, colortbl}
\usepackage{makecell}
\usepackage{multirow}
\usepackage{amsmath}
\usepackage{bm}
\usepackage[final]{pdfpages}

\usepackage{newfloat}
\usepackage{listings}
\DeclareCaptionStyle{ruled}{labelfont=normalfont,labelsep=colon,strut=off} 
\floatstyle{ruled}
\newfloat{listing}{tb}{lst}{}
\floatname{listing}{Listing}
\title{FatesGS: Fast and Accurate Sparse-View Surface Reconstruction using Gaussian Splatting with Depth-Feature Consistency}
\author {
    Han Huang\textsuperscript{\rm 1,\rm 2}\equalcontrib,
    Yulun Wu\textsuperscript{\rm 1, \rm 2}\equalcontrib,
    Chao Deng\textsuperscript{\rm 1, \rm 2},
    Ge Gao\textsuperscript{\rm 1, \rm 2}\thanks{Corresponding author.},
    Ming Gu\textsuperscript{\rm 1, \rm 2},
    Yu-Shen Liu\textsuperscript{\rm 2}
}
\affiliations {
    \textsuperscript{\rm 1}Beijing National Research Center for Information Science and Technology (BNRist), Tsinghua University, Beijing, China\\
    \textsuperscript{\rm 2}School of Software, Tsinghua University, Beijing, China\\
    \{h-huang20, wu-yl22, dengc23\}@mails.tsinghua.edu.cn, \{gaoge, guming, liuyushen\}@tsinghua.edu.cn
}

\usepackage{bibentry}

\begin{document}

\maketitle

\begin{abstract}
Recently, Gaussian Splatting has sparked a new trend in the field of computer vision. Apart from novel view synthesis, it has also been extended to the area of multi-view reconstruction. The latest methods facilitate complete, detailed surface reconstruction while ensuring fast training speed. However, these methods still require dense input views, and their output quality significantly degrades with sparse views. We observed that the Gaussian primitives tend to overfit the few training views, leading to noisy floaters and incomplete reconstruction surfaces. In this paper, we present an innovative sparse-view reconstruction framework that leverages intra-view depth and multi-view feature consistency to achieve remarkably accurate surface reconstruction. Specifically, we utilize monocular depth ranking information to supervise the consistency of depth distribution within patches and employ a smoothness loss to enhance the continuity of the distribution. To achieve finer surface reconstruction, we optimize the absolute position of depth through multi-view projection features. Extensive experiments on DTU and BlendedMVS demonstrate that our method outperforms state-of-the-art methods with a speedup of 60x to 200x, achieving swift and fine-grained mesh reconstruction without the need for costly pre-training.
\end{abstract}

%

\section{Introduction}
Reconstructing surfaces from multi-view images \cite{ramon2021h3d,chen2024vcr,zhang2024neural} is a long-standing task in 3D vision, graphics, and robotics. Multi-View Stereo \cite{depth2,mvsnet,depth3} is a traditional reconstruction method consisting of processes such as feature extraction, depth estimation, and depth fusion. This technique achieves favorable results with dense views, but struggles in sparse view reconstruction due to the lack of matching features.

Over the recent years, neural implicit reconstruction has rapidly progressed based on neural radiance fields \cite{nerf}. Some methods \cite{neus,volsdf,geoneus} employ neural rendering to optimize implicit geometry fields and color fields from multi-view images. They can achieve smooth and complete surfaces with implicit geometric representations. However, implicit geometric fields tend to overfit with a limited number of input views, leading to geometric collapse.
\begin{figure}[!t]
    \centering
    \includegraphics[width=\columnwidth]{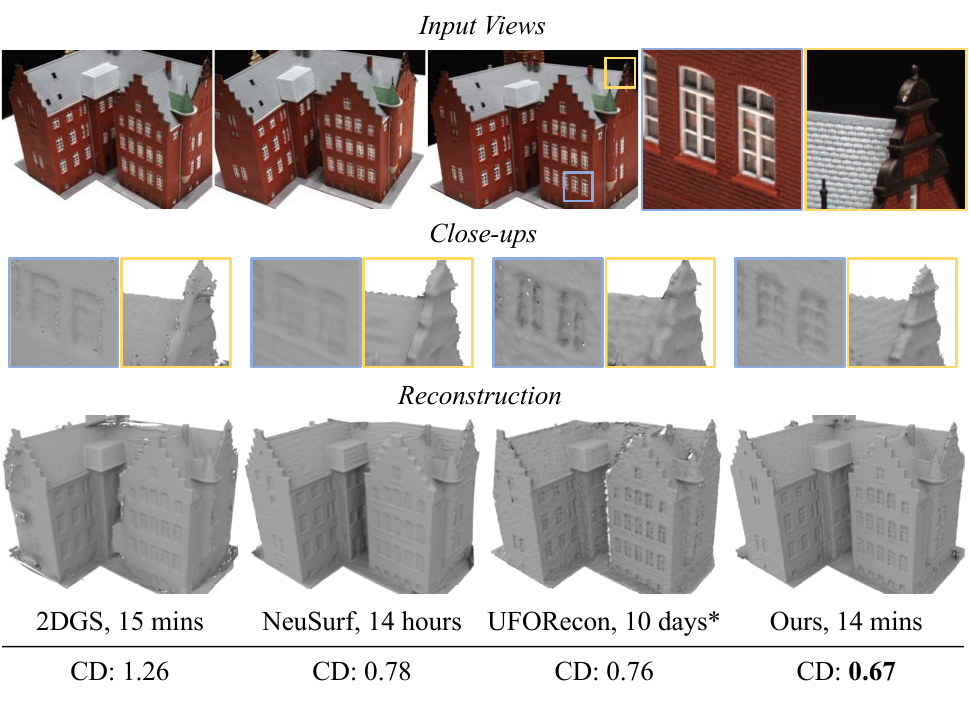}
    \caption{Surface reconstruction from 3-view images of DTU scan 24. The trendiest general method 2DGS \cite{2dgs} is fast but yields coarse results. The state-of-the-art per-scene optimization method, NeuSurf~\cite{huang2024neusurf}, and the generalization method, UFORecon~\cite{na2024uforecon}, produce suboptimal surfaces and require long training time. In contrast, our method (FatesGS) achieves swift and detailed reconstruction. *Pre-training time.} 
\label{fig:teaser}
\vspace{-0.5em}
\end{figure}

To address this issue, two types of neural implicit methods for sparse view reconstruction have been developed. The first type is a generalizable approach \cite{sparseneus,volrecon,na2024uforecon}, which is trained on large-scale datasets and subsequently applied to infer new scenes. The second type focuses on per-scene optimization \cite{monosdf,huang2024neusurf}, where no pre-training is needed, and the method directly fits different scenes. Although both types of methods achieve satisfactory geometric results, they require either several days of pre-training or optimization for several hours per scene, as shown in Figure \ref{fig:teaser}.

Lately, Gaussian Splatting \cite{3dgs} has been widely adopted for novel view synthesis due to its high rendering quality and fast training speed. However, 3D Gaussians lack the capability to represent scene geometry consistently, leading to imprecise surface reconstruction. To ensure the surface alignment property, some methods \cite{2dgs,gaussiansurfels,dnsplatter} modify the shape of Gaussian primitives and the splatting techniques. With depth maps fusion, the geometry of the object can be reconstructed completely and precisely. These methods retain the fast training speed of Gaussian Splatting in multi-view reconstruction. However, with fewer input views, geometric consistency decreases, leading to inaccurate Gaussian primitive localization and flawed depth rendering. This results in noisy and incomplete output meshes. 

In this paper, we present a novel sparse view reconstruction framework that leverages the efficient pipeline of Gaussian Splatting along with two consistency constraints to enhance both reconstruction speed and accuracy. Specifically, we transform the 3D ellipsoid Gaussian into a 2D ellipse Gaussian for more precise geometric representation and employ 2D Gaussian rendering to optimize the attributes of the Gaussian primitives. To mitigate local noise induced by overfitting, we segment the image into patches and regulate the ranking relationships within these patches using monocular depth information. Additionally, we introduce a smoothing loss to address abrupt depth changes in texture-less regions, thereby ensuring the continuity of the depth distribution. The intra-view depth consistency aided in achieving coarse reconstruction geometry, yet compromised numerous details. To resolve the issue of over-smoothing, we align the reprojection features of depth-rendered points to ensure precise multi-view feature consistency, which significantly enhances the quality of surface reconstruction. 

Our contributions are summarized as follows.
\begin{itemize}
    \item We propose FatesGS for sparse-view surface reconstruction, taking full advantage of the Gaussian Splatting pipeline. Compared with previous methods, our approach neither requires long-term per-scene optimization nor costly pre-training.
    \item We leverage intra-view depth consistency to facilitate the learning of coarse geometry. Furthermore, we optimize the multi-view feature consistency of depth-rendered points to enhance the learning of detailed geometry.
    \item We achieve state-of-the-art results in sparse view surface reconstruction under two distinct settings on the widely used DTU and BlendedMVS datasets.
\end{itemize}

\section{Related Works}
\subsection{Multi-View Stereo (MVS)}
In the field of 3D reconstruction, MVS methods have established themselves based on their scalability, robustness, and accuracy. Point clouds \cite{pointbased2,pointbased1}, depth maps \cite{depth1, depth2, depth3}, and voxel grids \cite{voxel1,surfacenet,voxel2} are used as 3D representations in MVS pipeline to accomplish geometry reconstruction. While these methods can achieve dense reconstruction, they often produce limited results in texture-less regions.
\subsection{Neural Implicit Reconstruction}
NeRF \cite{nerf} represents a scene as density and radiance fields, which are optimized using volumetric rendering. Inspired by this, NeuS \cite{neus}, VolSDF \cite{volsdf}, and subsequent optimization methods \cite{monosdf,geoneus,neuralwarp,li2023neuralangelo} transform signed distance function(SDF) into density, reconstructing multi-view images into implicit surfaces. However, these methods focus on dense view reconstruction, which places high demands on the input.

To enable sparse view reconstruction, both generalization and per-scene optimization methods have been proposed recently. The generalizable methods \cite{sparseneus,volrecon,c2f2neus,peng2023gens,liang2024retr,na2024uforecon} are trained on large-scale datasets and then generalized to new scenes. These methods require significant time on high-performance GPUs (usually several days) to learn the correspondence between 3D geometry and 2D views in advance. In contrast, per-scene optimization methods \cite{monosdf,divinet,wang2023sparsenerf,somraj2023vip,somraj2023simplenerf,huang2024neusurf} do not require training on large-scale datasets but instead directly fit the 3D geometry from the sparse images of a given scene. Due to the lack of learned correspondence, these methods often require several hours to fit from scratch.

\subsection{Gaussian Splatting}
3D Gaussian Splatting (3DGS) \cite{3dgs} represents the latest advancement in novel view synthesis, leveraging explicit Gaussian primitives for scene representation. By integrating a splatting-rendering pipeline, 3DGS maintains high-quality rendering while enabling real-time performance. However, 3DGS still requires dense view input and tends to overfit the training views when dealing with sparse input. To address this issue, some studies introduce monocular depth regularization \cite{fsgs,depthreggs,dngaussian,han2024binocular} for sparse views to constrain geometric relationships, thereby reducing Gaussian overfitting for high-quality rendering. 

Lately, to extend the advantages of Gaussian Splatting into the field of surface reconstruction, some work \cite{neusg, sugar,3dgsr} have enhanced surface representation by integrating regularization terms and Signed Distance Function (SDF) implicit fields at the cost of reduced training speed. 2DGS \cite{2dgs} and Gaussian Surfels \cite{gaussiansurfels} flatten the 3D ellipsoid into 2D ellipse to obtain more stable and consistent geometric surfaces. Although these methods achieve satisfactory results with dense views, they can only produce noisy and incomplete surfaces under sparse input.
\begin{figure*}
    \centering
    \includegraphics[width=\textwidth]{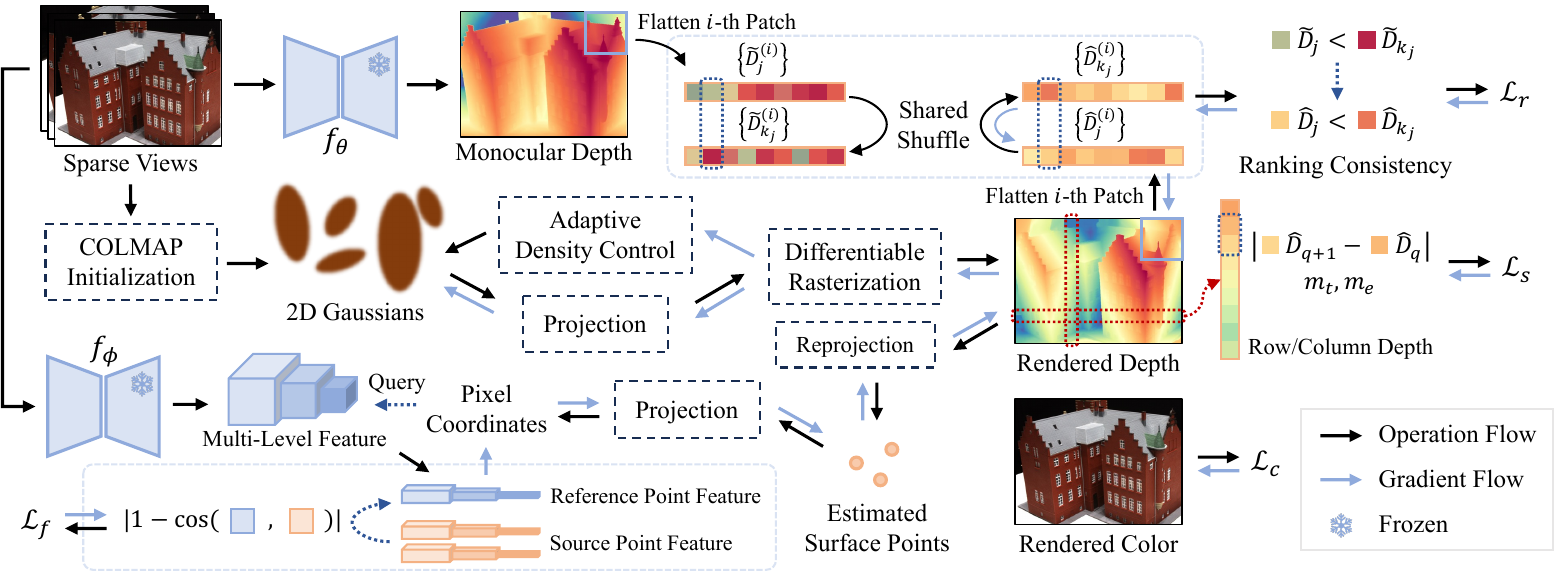}
    \caption{Overview of FatesGS. Starting with a set of sparse input views, we initialize 2D Gaussians using COLMAP and employ splatting to render RGB images and depth maps. To enhance the geometric learning process, we integrate ranking information from monocular depth estimation and apply depth smoothing to ensure intra-view depth consistency. To further refine the geometry, we align the multi-view features extracted by projecting estimated surface points onto the source images.}
    \label{fig:pipeline}
\end{figure*}

\section{Method}
Our goal is to reconstruct the high-quality geometry $\mathcal{S}$ of a scene from a collection of sparse-view images $\mathcal{I}=\{I_i \mid i \in 1,2 \ldots, {N}\}$, with poses $\mathcal{T}=\{T_i \mid i \in 1,2 \ldots, {N}\}$. In this paper, we propose \textit{FatesGS}, a Gaussian surface reconstruction approach with sparse views, as shown in Figure \ref{fig:pipeline}.

Since the Gaussian splatting process involves localized operations for fast rendering and optimization, it tends to produce floating artifacts and view misalignments when only a few views are provided \cite{sparse3dgs}. This results in the collapse of the learned geometry. Our motivation is to leverage intra-view depth consistency to prevent local noise for coarse geometry and multi-view feature alignment to maintain coherent observations for detailed geometry.

\subsection{Learning Multi-View Geometry by Gaussian Splatting}
3DGS \cite{3dgs} represents the scene as a series of 3D Gaussians. Each Gaussian can be defined by center position $\bm{\mu}$, scaling matrix $S$, rotation matrix $R$, opacity $o$, and SH coefficients. The view-dependent appearance can be rendered with local affine transformation \cite{volumesplatting} and alpha blending techniques. Although 3DGS can achieve good rendering results, the geometric results remain noisy.

Following the previous work \cite{2dgs,gaussiansurfels}, we flatten the 3D ellipsoid into 2D ellipse to enable the primitives to better cover the surface of objects. Scaling matrix $S$ and rotation matrix $R$ can be expressed as $S=(s_1,s_2)$, $R=(\bm{t}_1,\bm{t}_2,\bm{t}_1\times \bm{t}_2)$. 
Then the 2D ellipse can be defined within a local tangent plane in world space as:
\begin{equation}
P(u,v) = \bm{\mu} + s_1 \bm{t}_1 u + s_2 \bm{t}_2 v .
\label{eq:plane-to-world}
\end{equation}
For the point $\bm{u} = (u, v)$ within the $uv$ plane, its corresponding 2D Gaussian value can be determined using the standard Gaussian function:
\begin{equation}
\mathcal{G}(\bm{u}) = \exp\left(-\frac{u^2+v^2}{2}\right).
\label{eq:gaussian-2d}
\end{equation} 
During the scene optimization process, the parameters of the 2D Gaussian primitives are all designed to be learnable. The view-dependent color $\bm{c}$ is obtained through spherical harmonic (SH) coefficients. For Gaussian rasterization, 2D Gaussians are depth-sorted and then integrated into an image with alpha blending from front to back.
Given a pixel from one image, the rendered color $\hat{\bm{C}}(\bm{r})$ of a homogeneous ray $\bm{r}$ emitted from the camera can be expressed as:
\begin{equation}
\hat{\bm{C}}(\bm{r}) = \sum_{i=1} \bm{c}_i \omega_i ,
\end{equation}
\begin{equation}
\omega_i = o_i \mathcal{G}_i\left(\bm{u}(\bm{r})\right) \prod_{j=1}^{i-1} \left(1 - o_j \mathcal{G}_j(\bm{u}(\bm{r}))\right) ,
\end{equation}
where $\bm{c}_i$ is the $i$-th view-dependent color, $\omega_i$ is blending weight of the $i$-th intersection.

Similarly, the rendered depth $\hat{D}(\bm{r})$ for the homogeneous ray $\bm{r}$ can be accumulated by alpha blending as:
\begin{equation}
\hat{D}(\bm{r}) = \frac{\sum_{i=1} \omega_i d_i}{\sum_{i=1} \omega_i + \epsilon}.
\end{equation}
Following \cite{2dgs}, the $i$-th intersection depth $d_i$ is obtained by the ray-splat intersection algorithm.
\subsection{Intra-View Depth Consistency}
Since Gaussian Splatting lacks the concept of geometric fields, surface reconstruction relies on rendered depth extraction. Direct depth optimization seems to avoid overfitting and address geometric noise effectively. Employing absolute scaling for monocular depth to supervise rendered depth \cite{monosdf,sparsegs} and enhancing the correlation between monocular and rendered depth \cite{fsgs} are regarded as effective depth regularization techniques. However, it has been proven that these strategies might result in a noisy distribution of Gaussian primitives \cite{sun2024uncertainty}.
To avoid geometric collapse caused by hard constraints, we utilized monocular depth information to maintain the ranking consistency of local rendering depth. Since long-range depth ambiguity may exist in monocular depth, we performed local depth information distillation on a patch-by-patch basis.

Specifically, We divide the image $I$ into patches, each of size $M \times M$. The $i$-th patch $\mathcal{P}_i$ is represented as a list of pixels: 
\begin{equation}
\mathcal{P}_i=\left\{\bm{p}^{(i)}_{j}\ \Big|\ j\in 1,\dots,{M^2}\right\}.
\end{equation}
To simplify, the pixels in the patch are shuffled, denoted as:
\begin{equation}
\mathcal{P}_i^\prime=\mathrm{shuffle}\left(\mathcal{P}_i\right) =\left\{\bm{p}_{k_j}^{(i)}\ \Big|\ j\in1,\dots,{M^2}\right\}.
\end{equation}
For each pixel in $\mathcal{P}_i$ and $\mathcal{P}_i^\prime$, we obtain its rendered depth $\hat{D}$ and monocular depth $\widetilde{D}$. A patch-based depth ranking loss is then expressed as:
\begin{equation}
\mathcal{L}_{r}=\sum_{i,j}\sigma\left(\mathrm{sgn}\left(\widetilde{D}^{(i)}_{k_j}-\widetilde{D}^{(i)}_{j}\right)\cdot\left(\hat{D}^{(i)}_{j}-\hat{D}^{(i)}_{k_j}\right)+m\right),
\end{equation}
where $\sigma(\cdot)$ represents the ReLU function, and $m$ is a small positive threshold. 

The patch-based depth ranking loss ensures the overall distribution consistency of Gaussian primitives. However, noisy primitives still exist in texture-less areas, resulting in abrupt depth changes. Therefore, we propose a smoothing loss for the depth of adjacent pixels to enhance the distribution continuity of the reconstructed surface:

\begin{equation}
\mathcal{L}_{s}=\sum_{i,j,k}\sum_{\left|\widetilde{D}_k-\widetilde{D}_{(i,j)}\right|<m_e}\sigma\left(\left|\hat{D}_k-\hat{D}_{(i,j)}\right|-m_t\right).
\end{equation}
Here, $\hat{D}_{(i,j)}$ denotes the rendered depth value of the pixel at $i$-th row and $j$-th column within the whole image. Small positive thresholds $m_e$ and $m_t$ are utilized to recognize edges and avoid over-smoothing. $k\in\{(i+1,j),(i,j+1)\}$.

\subsection{Multi-View Feature Alignment}
Intra-view depth consistency helps maintain the overall shape and structure of the reconstructed object. While ranking and smoothing are effective in reducing artifacts and preserving the coarse geometry, they fall short in refining the finer details of the reconstruction. Multi-view geometry may present a reliable solution. The traditional Multi-View Stereo (MVS) reconstruction pipeline typically employs photometric consistency across multiple views to refine the surface. Inspired by that, a straightforward idea is to project the 3D points corresponding to the depth of each view onto other views and then compute the color difference on the projected views. 

However, due to the influence of lighting, the colors may differ across different viewpoints \cite{zhan2018unsupervised}. When there are only a few input views, the number of reference views for projection is limited, and the spacing between views is greater compared to dense views. As a result, the influence of lighting on the color of surface points becomes more pronounced. To resolve these issues, we have designed a multi-level feature projection loss.

Let $I^{(l)}_i$ denote the image whose resolution is downscaled by a scale factor $l$ from the original image $I_i$, the image set of the downscaled images then can be marked as
\begin{equation}
\mathcal{I}^{(l)}=\left\{I^{(l)}_{i}\ \big|\ i\in1,2,\ldots,N\right\},\ l\in1,2,\ldots,2^L.
\end{equation}

Multi-view feature at single level $l$ can be calculated using a frozen feature extraction network $f_{\phi}$:
\begin{equation}
\mathcal{F}^{(l)}=f_{\phi}(\mathcal{I}^{(l)})=\left\{\bm{F}^{(l)}_{i}\ \big|\ i\in1,2,\ldots,N\right\}.
\end{equation}

Let $I_r$, $I_{s}$ denote the reference view image and one of its source view images, respectively. For a pixel $\bm{p}_{r,i}$ of $I_r$, with its rendered depth $\hat{D}_{r,i}$, we can calculate the corresponding spatial point $\bm{x}_{r,i}$ and its projected pixel coordinate $\bm{p}_{s,i}$ to the source view $I_{s}$ by
\begin{equation}
\bm{x}_{r,i}=\bm{o}_{r}+\hat{D}_{r,i}\cdot\bm{d}_{r,i},
\end{equation}
\begin{equation}
\bm{p}_{s,i} = KP_{{s}}^{-1}\bm{x}_{r,i},
\end{equation}
where $K$ and $P_{s}$ represent the intrinsic matrix and camera pose of the source view image $I_s$. $\bm{d}_{r,i}$ is the normalized direction vector of the ray omitted form $\bm{o}_r$ passing through $\bm{p}_{r,i}$. Then the feature loss can be acquired by
\begin{equation}
\mathcal{L}_{f} = \sum_{s,i,l}\frac{1}{l}\cdot v_{r,s,i}\left|{1}-\cos\left(\bm{F}^{(l)}_{r}\left(\bm{p}_{r,i}\right),\bm{F}^{(l)}_{{s}}\left(\bm{p}_{{s},i}\right)\right)\right|.
\label{eqn:feature_loss}
\end{equation}
Since surface points may be occluded when projected onto the source views. We design visibility item $v_{r,s,i}$, which indicates the visibility of $\bm{x}_{r,i}$ from the viewpoint $\bm{o}_{s}$. For spatial points along the ray $\bm{r}_{s,i}$ which is emitted from $\bm{o}_{s}$ and passes through $\bm{p}_{s,i}$, only the nearest one is considered visible and its visibility item is set as 1, while the others are set as 0. The process can be expressed as
\begin{equation}
\begin{split}
&v_{r,s,i}=\left[i=\mathop{\arg\min}\limits_{t}\left(\left\Vert\bm{x}_{r,t}-\bm{o}_{s}\right\Vert\right)\right],\\
&\mathrm{where}\ \ t\in\left\{t\ \Big|\ \bm{p}_{s,i}=KP_{{s}}^{-1}\bm{x}_{r,t}\right\}.
\end{split}
\end{equation}
$[\cdot]$ represent the Iverson Bracket.

\subsection{Loss Functions}
\begin{table*}[!ht]
    \centering
    \fontsize{8pt}{10pt}\selectfont
        \begin{tabular}{l|*{15}{c}|c}
            \toprule
            Scan ID & 24 & 37 & 40 & 55 & 63 & 65 & 69 & 83 & 97 & 105 & 106 & 110 & 114 & 118 & 122 & Mean \\
            \midrule
            \midrule
            {COLMAP} & 0.90 & 2.89 & 1.63 & 1.08 & 2.18 & 1.94 & 1.61 & 1.30 & 2.34 & 1.28 & 1.10 & 1.42 & 0.76 & 1.17 & 1.14 & 1.52 \\
            TransMVSNet & 1.07 & 3.14 & 2.39 & 1.30 & 1.35 & 1.61 & {0.73} & 1.60 & 1.15 & 0.94 & 1.34 & {\textbf{0.46}} & 0.60 & 1.20 & 1.46 & 1.35 \\
            \cmidrule(l{0.5em}r{0.5em}){1-17}
            {SparseNeuS$_{ft}$} & 1.29 & {2.27} & 1.57 & 0.88 & 1.61 & 1.86 & 1.06 & 1.27 & 1.42 & 1.07 & 0.99 & 0.87 & 0.54 & 1.15 & 1.18 & 1.27 \\ 
            {VolRecon} & 1.20 & 2.59 & 1.56 & 1.08 & 1.43 & 1.92 & 1.11 & 1.48 & 1.42 & 1.05 & 1.19 & 1.38 & 0.74 & 1.23 & 1.27 & 1.38 \\
            ReTR & 1.05 & {2.31} & 1.44 & 0.98 & 1.18 & {1.52} & 0.88 & 1.35 & 1.30 & 0.87 & 1.07 & 0.77 & 0.59 & 1.05 & 1.12 & 1.17 \\
            C2F2NeuS & 1.12 & 2.42 & {1.40} & {\textbf{0.75}} & 1.41 & 1.77 & 0.85 & {1.16} & 1.26 & 0.76 & 0.91 & 0.60 & {0.46} & 0.88 & {0.92} & 1.11 \\
            GenS$_{ft}$ & 0.91 & 2.33 & 1.46 & {\textbf{0.75}} & {\textbf{1.02}} & 1.58 & {0.74} & {1.16} & {1.05} & 0.77 & {0.88} & {0.56} & 0.49 & {\underline{0.78}} & 0.93 & 1.03 \\
            UFORecon & \underline{0.76} & \underline{2.05} & \underline{1.31} & 0.82 & {1.12} & {\textbf{1.18}} & {0.74} & 1.17 & 1.11 & {0.71} & {0.88} & 0.58 & 0.54 & 0.86 & 0.99 & \underline{0.99} \\
            \cmidrule(l{0.5em}r{0.5em}){1-17}
            NeuS & 4.57 & 4.49 & 3.97 & 4.32 & 4.63 & 1.95 & 4.68 & 3.83 & 4.15 & 2.50 & 1.52 & 6.47 & 1.26 & 5.57 & 6.11 & 4.00  \\
            VolSDF & 4.03 & 4.21 & 6.12 & 0.91 & 8.24 & 1.73 & 2.74 & 1.82 & 5.14 & 3.09 & 2.08 & 4.81 & 0.60 & 3.51 & 2.18 & 3.41 \\
            MonoSDF & 2.85 & 3.91 & 2.26 & 1.22 & 3.37 & 1.95 & 1.95 & 5.53 & 5.77 & 1.10 & 5.99 & 2.28 & 0.65 & 2.65 & 2.44 & 2.93 \\
            NeuSurf & {0.78} & 2.35 & 1.55 & \textbf{0.75} & \underline{1.04} & 1.68 & {\textbf{0.60}} & \underline{1.14} & {\textbf{0.98}} & \underline{0.70} & {\textbf{0.74}} & {0.49} & {\textbf{0.39}} & {\textbf{0.75}} & {\textbf{0.86}} & \underline{0.99} \\
            \cmidrule(l{0.5em}r{0.5em}){1-17}
            3DGS & 3.38 & 4.19 & 2.99 & 1.76 & 3.38 & 3.80 & 5.21 & 2.91 & 4.29 & 3.18 & 3.23 & 5.18 & 2.78 & 3.48 & 3.32 & 3.54 \\
            Gaussian Surfels & 3.56 & 5.42 & 3.95 & 3.68 & 4.61 & 2.72 & 4.42 & 5.22 & 4.71 & 3.46 & 4.07 & 5.42 & 2.44 & 3.27 & 4.00 & 4.06 \\
            2DGS & 1.26 & 2.95 & 1.73 & 0.96 & 1.68 & 1.97 & 1.58 & 1.87 & 2.50 & 1.02 & 1.93 & 1.91 & 0.72 & 1.85 & 1.37 & 1.69 \\
            \cmidrule(l{0.5em}r{0.5em}){1-17}
            \textbf{Ours} & {\textbf{0.67}} & {\textbf{1.94}} & {\textbf{1.17}} & 0.77 & 1.28 & \underline{1.23} & \underline{0.63} & {\textbf{1.05}} & {\textbf{0.98}} & {\textbf{0.69}} & \underline{0.75} & \underline{0.48} & \underline{0.41} & \underline{0.78} & \underline{0.90} & \textbf{0.92} \\
            \bottomrule
        \end{tabular}
    \caption{The quantitative comparison results of Chamfer Distance (CD$\downarrow$) on DTU dataset (large-overlap setting). In this table, the best results are in bold, the second best are underlined.}
    \label{tab:cd_results}
\end{table*}
The overall loss functions are defined as follows:
\begin{equation}
\mathcal{L} = \mathcal{L}_{c} +\lambda_{1} \mathcal{L}_{r} 
+\lambda_{2} \mathcal{L}_{s} +\lambda_{3} \mathcal{L}_{f} +\lambda_{4} \mathcal{L}_{d}+\lambda_{5} \mathcal{L}_{n},
\end{equation}
where $\mathcal{L}_{r}$ and $\mathcal{L}_{s}$ represent the ranking and smoothing losses from intra-view depth consistency, respectively, and $\mathcal{L}_{f}$ denotes the multi-view feature loss.

According to 3DGS \cite{3dgs}, the $\mathcal{L}_1$ loss and $\mathcal{L}_{D-SSIM}$ loss are utilized for color supervision $\mathcal{L}_{c}$. This can be formulated as follows, with $\lambda=0.2$:
\begin{equation}
\mathcal{L}_{c} = (1-\lambda)\mathcal{L}_1 + \lambda \mathcal{L}_{D-SSIM}.
\end{equation}
As with 2DGS \cite{2dgs}, depth distortion loss and normal consistency loss are used as regularization terms to optimize surface geometry.
\begin{equation}
\mathcal{L}_{d} = \sum_{i,j}\omega_i\omega_j\left| d_i-d_j \right|, \quad
\mathcal{L}_{n} = \sum_{i}\omega_i(1-\bm{n}^T_i\bm{N}) .
\end{equation}
Here, $\omega$ and $d$ are computed during the Gaussian Splatting process, $\bm{n}^T_i$ represents the estimated normal near the depth point, and $\bm{N}$ is the estimated normal near the depth point.

\section{Experiments and Analysis}
To demonstrate the effectiveness and generalization performance of our approach, we compare our evaluation results with previous state-of-the-art methods in terms of reconstruction accuracy and training efficiency. Additionally, we provide a detailed ablation study and analysis to validate the efficacy of each component of our proposed method.

\subsection{Experimental Settings}
\subsubsection{Datasets.}
We evaluate our approach on DTU dataset \cite{dtu}, which is extensively utilized in previous surface reconstruction research. DTU comprises 15 scenes, each with 49 or 69 images at a resolution of 1600 $\times$ 1200. We follow the previous work \cite{huang2024neusurf} to train and evaluate the model on 3 views of both the large-overlap (SparseNeuS) setting and the little-overlap (PixelNeRF) setting. The images are downscaled into 800 $\times$ 600 pixels during training procedure, following \cite{2dgs}. To assess generalization performance, we further test our method on BlendedMVS dataset \cite{blendedmvs} with randomly selected 3 input views per scene at a resolution of 768 $\times$ 576. Consistent with sparse-view settings from previous works, the camera poses are assumed to be known.

\subsubsection{Baselines.}
We compare our approach with abundant SOTA methods of various categories. \textbf{i.} MVS methods: COLMAP \cite{colmap} and TransMVSNet \cite{ding2022transmvsnet}. \textbf{ii.} Generalizable sparse-view neural implicit reconstruction methods: SparseNeuS \cite{sparseneus}, VolRecon \cite{volrecon}, ReTR \cite{liang2024retr}, C2F2NeuS \cite{c2f2neus}, GenS \cite{peng2023gens} and UFORecon \cite{na2024uforecon}. \textbf{iii.} Per-scene optimization neural implicit methods: NeuS \cite{neus}, VolSDF \cite{volsdf}, MonoSDF \cite{monosdf} and NeuSurf \cite{huang2024neusurf}. \textbf{iv.} Gaussian splatting based methods: 3DGS \cite{3dgs}, Gaussian Surfels \cite{gaussiansurfels} and 2DGS \cite{2dgs}. For a fair comparison, we initialize 3DGS and 2DGS with the same point clouds used in our method. We also adopt the same TSDF depth fusion approach as ours for 3DGS to extract meshes.
\begin{figure}[!t]
    \centering
    \includegraphics[width=\columnwidth]{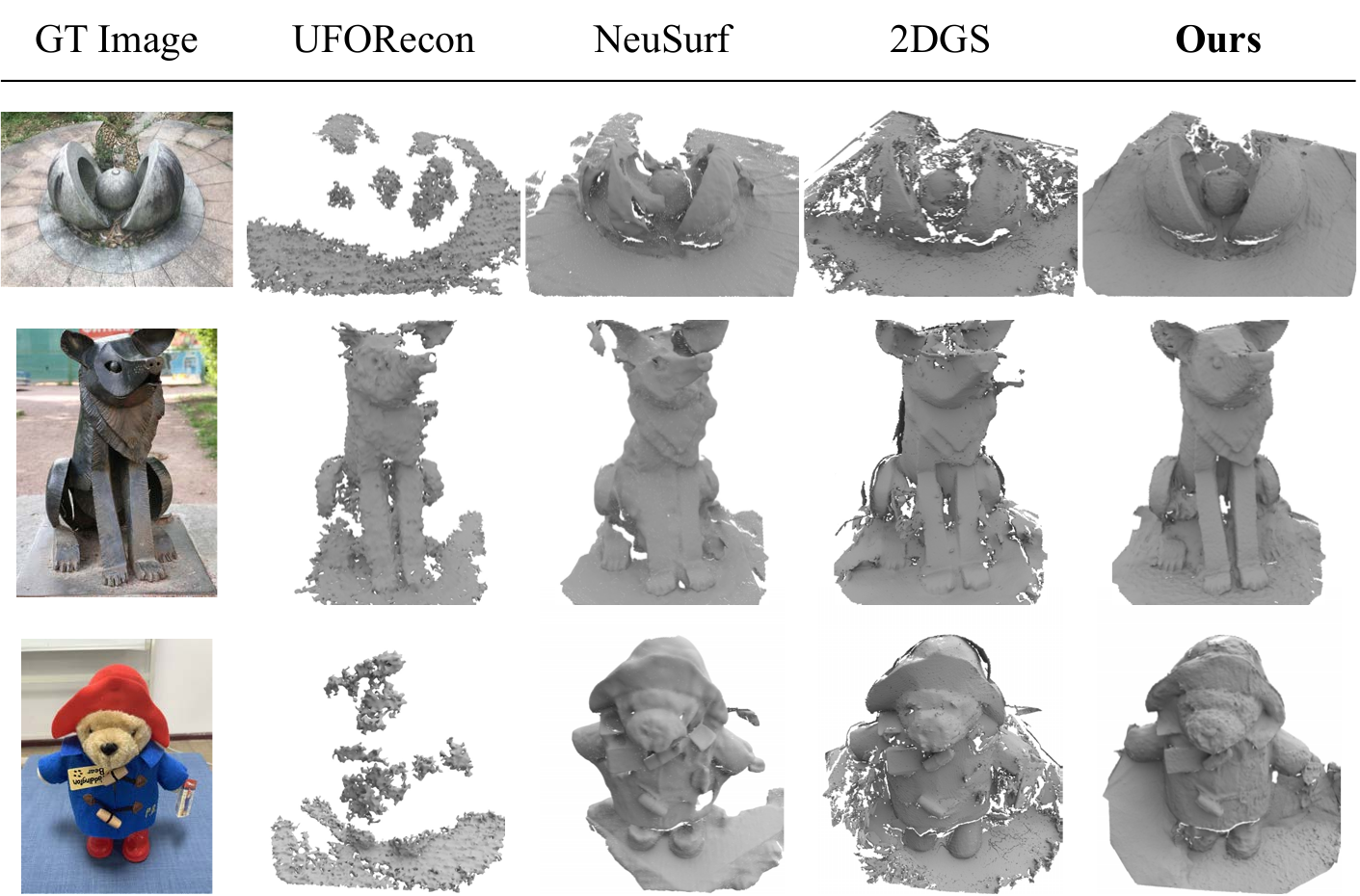}
    \caption{Visual comparison of 3-view reconstruction on BlendedMVS dataset.}%
\label{fig:bmvs}
\end{figure}
\begin{figure*}[!ht]
    \centering
    \includegraphics[width=0.95\textwidth]{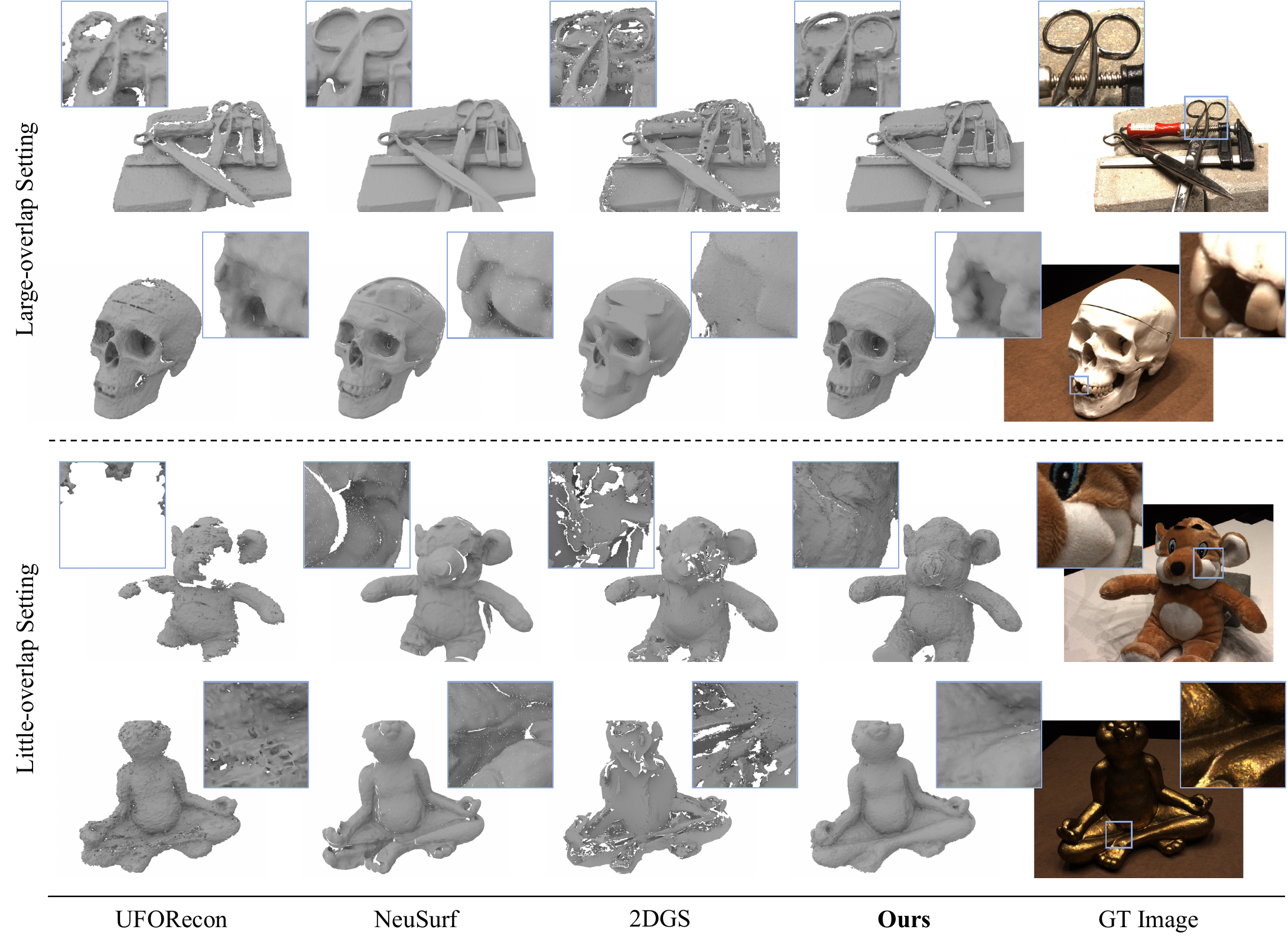}
    \caption{Qualitative comparison of reconstruction results on the DTU with different sparse settings.}
    \label{fig:comparison}
\end{figure*}

\subsubsection{Implementation Details.}
Following previous research, we use COLMAP \cite{colmap} for Gaussians initialization. Our framework is built upon 2DGS \cite{2dgs} and 3DGS \cite{3dgs}. We adopt Vis-MVSNet \cite{vismvsnet} as the feature extraction network $f_{\phi}$ and Marigold \cite{marigold} as the monocular depth estimation model $f_{\theta}$. All experiments presented in this paper are conducted on a single NVIDIA RTX 3090 GPU.

\subsection{Comparisons}
\subsubsection{Sparse View Reconstruction.}
The quantitative results of geometry reconstruction from sparse input views on the DTU dataset (large-overlap setting) are presented in Table \ref{tab:cd_results}. Additional experimental results (e.g., little-overlap setting) are presented in the supplementary materials. Our method achieves the best mean Chamfer Distance (CD) performance across 15 scenes compared to others. As illustrated in Figure \ref{fig:comparison}, our approach achieves more comprehensive global geometry and preserves finer details. This highlights our method's superior capability in multi-view feature extraction.  Moreover, in contrast to NeuSurf, our method successfully avoids over-smoothing of the geometric surfaces.

Reconstruction results on BlendedMVS are shown in Figure \ref{fig:bmvs}. Our method exhibits consistent and stable performance across datasets with the same set of hyperparameters. In contrast, UFORecon, which is currently the most recent generalizable method, has not undergone extensive training on this dataset, resulting in significant reconstruction defects and noise. NeuSurf, being the latest per-scene optimization method, produces surfaces in the SDF field that are overly smooth, leading to a loss of local texture details. 2DGS, a leading Gaussian splatting surface reconstruction method, struggles with sparse image coverage. Insufficient geometric consistency can lead to flawed depth rendering and suboptimal reconstruction results.

\begin{table}[!ht]
    \centering
    \small
    \begin{tabular}{l|c|c|c}
    \toprule
        \multirow{2.5}{*}{Method} & \multicolumn{2}{c|}{Training Time} & \multirow{2.5}{*}{\makecell[c]{GPU Mem.}}\\
        \cmidrule{2-3}
        & \makecell[c]{Pre-Training} & \makecell[c]{Per-Scene} & \\
        \midrule
        SparseNeuS$_{ft}$ & 2.5 days & 19 mins & 7 GB \\
        GenS$_{ft}$ & $\sim$ 1 day$^*$ & 25 mins & 19 GB$^*$ \\
        VolRecon & $\sim$ 2 days & \multirow{3}{*}{-} & 17 GB \\
        ReTR & $\sim$ 3 days &  & 22 GB \\
        UFORecon & $\sim$ 10 days & & 23 GB \\
        MonoSDF & \multirow{3}{*}{-} & 6 hours & 14 GB \\
        NeuSurf &  & 14 hours & 8 GB \\
        \textbf{Ours} &  & \textbf{14 mins} & \textbf{4 GB} \\
    \bottomrule
    \end{tabular}
    \caption{Comparison with the efficiency of sparse-view reconstruction methods. The listed GPU memory values are approximate maximum occupancies during training. $^*$We used 2 NVIDIA RTX 3090 GPUs for GenS pre-training.}
    \label{tab:efficiency}
\end{table}
\subsubsection{Efficiency.}
We conduct an efficiency study on all specialized sparse-view reconstruction methods using the DTU SparseNeuS 3-view setting, as detailed in Table \ref{tab:efficiency}. The presented results are obtained from tests conducted on a single NVIDIA RTX 3090 GPU. To ensure a fair comparison, all models are configured with settings optimized for peak performance. In previous methods, generalizable approaches require extensive pre-training, often taking several days. Per-scene optimization methods, on the other hand, need several hours of training for each scene. In contrast, our method completes training in just a few minutes and uses significantly less GPU memory.

\subsubsection{Depth Prediction.}
We trained our model using three seen views and tested it on the same three views, along with three additional unseen views. We then calculated the error with the ground truth depth, and compared the results with methods Marigold \cite{marigold} and 2DGS \cite{2dgs}, as shown in Table \ref{tab:depth_eval}. Marigold is a universal method for monocular depth prediction, limited to predicting relative depth. To facilitate comparison, we rescale the predicted results to real-world dimensions using the ground truth depth. The results demonstrate that our method significantly outperforms both the 2D Gaussian Splatting (2DGS) backbone and the prior method, Marigold, in depth prediction. Our approach effectively integrates monocular depth information with the Gaussian splatting pipeline, leading to more consistent and accurate multi-view depth learning.
\begin{table}[!h]
    \centering
    \small
    \renewcommand\arraystretch{1.2}
    \begin{tabular}{l|ccc}
    \toprule
        Method & Marigold & 2DGS & Ours \\
        \midrule
        $<1\uparrow$ & 5.01 / 4.40 & 35.74 / 31.24 & \textbf{77.35} / \textbf{73.52} \\
        $<2\uparrow$ & 9.98 / 8.84 & 57.26 / 50.71 & \textbf{89.58} / \textbf{87.38} \\
        $<4\uparrow$ & 19.55 / 17.82 & 73.78 / 66.92 & \textbf{94.34} / \textbf{92.69} \\
        Abs. $\downarrow$ & 15.58 / 15.12 & 7.46 / 18.07 & \textbf{2.41} / \textbf{3.43} \\
        Rel. $\downarrow$ & 2.37 / 2.31 & 1.05 / 2.66 & \textbf{0.33} / \textbf{0.49} \\
    \bottomrule
    \end{tabular}
    \caption{Depth map evaluation results on DTU (seen / unseen). The result of mean absolute error (Abs.) is in millimeters. The result of threshold percentage ($<1\mathrm{mm},<2\mathrm{mm}$ and $<4\mathrm{mm}$) and mean absolute relative error (Rel.) are in percentage (\%). The best results are highlighted in bold.}
    \label{tab:depth_eval}
\end{table}

\subsection{Ablation Study}
\subsubsection{The Proposed Components.}
\begin{figure}[!ht]
    \centering
    \includegraphics[width=\columnwidth]{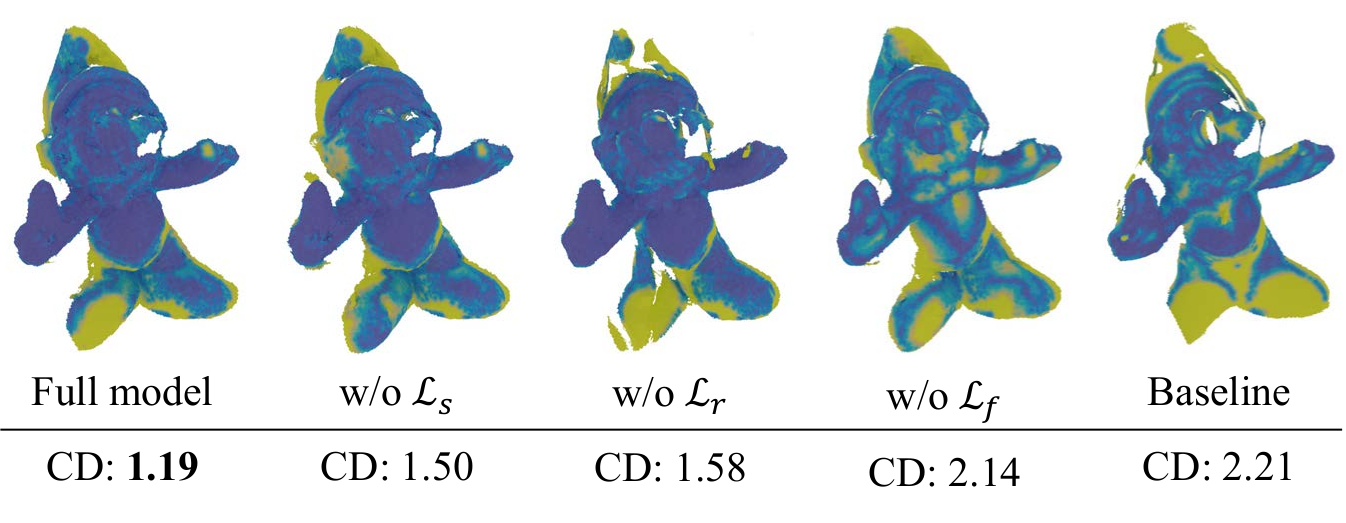}
    \caption{Visual comparison of ablation study on DTU scan 83. The transition of the error maps from blue to yellow indicates larger reconstruction errors.}
\label{fig:ablation}
\end{figure}

To demonstrate the effectiveness and necessity of each proposed component, we isolate individual design choices and measure their impact on reconstruction quality. Our experiments are conducted on the DTU dataset using the little-overlap setting, maintaining the same hyperparameters as in the main experiment. The mean Chamfer Distance (CD) values of all 15 scenes are reported in Table \ref{tab:ablation}. Furthermore, the ablation results for scan 83 are visualized in Figure \ref{fig:ablation}.
\begin{table}[h]
    \centering
    \small
    \begin{tabular}{cccccc}
    \toprule
        $\mathcal{L}_{r}$ & $\mathcal{L}_{s}$ & $\mathcal{L}_{f}$ & Mean CD$\downarrow$\\
        \midrule
        & & & 2.47 \\
        \raisebox{0.6ex}{\scalebox{0.7}{$\sqrt{}$}} & \raisebox{0.6ex}{\scalebox{0.7}{$\sqrt{}$}} &  & 2.56 \\
        \raisebox{0.6ex}{\scalebox{0.7}{$\sqrt{}$}} &  & \raisebox{0.6ex}{\scalebox{0.7}{$\sqrt{}$}} & 1.56 \\
         & \raisebox{0.6ex}{\scalebox{0.7}{$\sqrt{}$}} & \raisebox{0.6ex}{\scalebox{0.7}{$\sqrt{}$}} & 1.62 \\
        \raisebox{0.6ex}{\scalebox{0.7}{$\sqrt{}$}} & \raisebox{0.6ex}{\scalebox{0.7}{$\sqrt{}$}} & \raisebox{0.6ex}{\scalebox{0.7}{$\sqrt{}$}} & \textbf{1.37} \\
    \bottomrule
    \end{tabular}
    \caption{Comparison of reconstruction from the ablation study for the little-overlap setting on the DTU dataset.}
    \label{tab:ablation}
\end{table}
Removing each of the proposed optimization losses results in varying degrees of performance decline, demonstrating the effectiveness of each component. Notably, the model with only the intra-view depth ranking loss ($\mathcal{L}_{r}$) and the smoothing loss ($\mathcal{L}_{s}$) performs worse than the baseline model, which does not include any of the three losses. This indicates that the contributions of the three optimization losses to the full model are neither isolated nor merely additive. As shown in Figure \ref{fig:ablation}, $\mathcal{L}_{r}$ and $\mathcal{L}_{s}$ provide globally complete and coarsely correct geometric guidance. However, they cannot ensure local details due to the lack of absolute scale information. After incorporating feature loss ($\mathcal{L}_{f}$), we observe that the reconstructed surface details are significantly enhanced, effectively avoiding excessive smoothing.

\subsubsection{The Number of Training Views.}
To validate the impact of image counts on our proposed method, we varied the number of views, and the results are summarized in Table \ref{tab:more_views}. As the number of images increases, the reconstruction quality improves progressively. Incorporating additional views can enhance multi-view consistency, ensure stable reconstruction results, and prevent overfitting.
\begin{table}[!ht]
    \centering
    \small
    \begin{tabular}{lcccc}
    \toprule
        Number of Views & 3 & 6 & 9 & Full \\
        \midrule
        Mean CD$\downarrow$ & 0.92 & 0.85 & 0.79 & \textbf{0.61}\\
    \bottomrule
    \end{tabular}
    \caption{Ablation study of number of views on DTU dataset. The best result is highlighted in bold.}
    \label{tab:more_views}
\end{table}

\section{Conclusion}
In this paper, we present FatesGS, a novel method for sparse view surface reconstruction utilizing a Gaussian Splatting pipeline. To combat geometric collapse caused by overfitting in sparse views, we enhance the learning of coarse geometry through intra-view depth consistency. For finer geometric details, we optimize multi-view feature consistency. Our method is robust across various sparse settings and does not require large-scale training. Unlike previous methods, our approach eliminates the need for long-term per-scene optimization and expensive in-domain prior training. We demonstrate state-of-the-art results in sparse view surface reconstruction under two distinct settings, validated on the widely used DTU and BlendedMVS datasets.

\section{Acknowledgments}
The corresponding author is Ge Gao. This work was supported by Beijing Science and Technology Program (Z231100001723014).

\bibliography{aaai25}


\section*{Supplementary material}

# Supplementary Material for *FatesGS: Fast and Accurate Sparse-View Surface Reconstruction using Gaussian Splatting with Depth-Feature Consistency*

In this supplementary document, we provide additional framework and implementation details, and explain the evaluation metrics in Section **A**. In Section **B**, we present further validation experiments to demonstrate the superiority of our method in dense-view reconstruction, sparse-view reconstruction under various settings, and novel view synthesis from sparse views. We also compare our results with those obtained from direct monocular depth fusion. Section **C** provides an evaluation of our method under different conditions by conducting additional ablation studies. These studies focus on comparing intra-view depth supervision approaches and assessing the impact of multi-view feature consistency on surface reconstruction. In Section **D**, we discuss concurrent works and highlight the differences with our method through experimental results. In Section **E**, we discussed the applications of sparse reconstruction in other fields and explored potential avenues for future research. Section **F** addresses the potential negative social impacts of our research. Finally, Section **G** provides comprehensive visualized experimental results.

## A. Details of Framework and Implementation

### Framework

In the main paper, we introduce a novel framework, FatesGS, for surface reconstruction from sparse-view images. While FatesGS shares a foundational pipeline with 2DGS (Huang et al. 2024), it incorporates two primary constraints tailored for sparse view reconstruction. The core concept is to utilize intra-view depth consistency to mitigate local noise in coarse geometry, and multi-view feature alignment to ensure coherent observations in detailed geometry.

### Implementation

We have implemented the FatesGS based on both 2DGS and 3DGS. Specifically, we employed an adaptive density control strategy, similar to that used in 3DGS, to enhance the placement of Gaussian primitives. For the proposed loss functions, the loss weights $\lambda_1, \lambda_2, \lambda_3, \lambda_4$ and $\lambda_5$ are set to $10, 1, 1.5, 10^4$ and $0.05$, respectively. The feature levels $l = 1, 2$. Margins $m, m_e$ and $m_t$ are specified as $10^{-4}, 10^{-2}$ and $10^{-4}$. Experiments were carried out utilizing an NVIDIA RTX 3090 GPU alongside two Intel Xeon C6226R CPUs. The software setup operated on Ubuntu 18.04, with PyTorch version 2.0.0 and CUDA version 11.8.

### Evaluation Metrics

For quantitative evaluation of surface reconstruction, we follow previous methods and use L2 Chamfer Distance (CD) as the metric, which is calculated by

$$\text{CD}(\mathcal{S}_1, \mathcal{S}_2) = \frac{1}{2|\mathcal{S}_1|} \sum_{\boldsymbol{x} \in \mathcal{S}_1} \min_{\boldsymbol{y} \in \mathcal{S}_2} \|\boldsymbol{x} - \boldsymbol{y}\|_2 + \frac{1}{2|\mathcal{S}_2|} \sum_{\boldsymbol{y} \in \mathcal{S}_2} \min_{\boldsymbol{x} \in \mathcal{S}_1} \|\boldsymbol{x} - \boldsymbol{y}\|_2. \tag{1}$$

Here, $\mathcal{S}_1$ and $\mathcal{S}_2$ represent the sampled point cloud from the reconstructed mesh and the GT point cloud, respectively.

For depth prediction, we follow VolRecon (Ren et al. 2023) to take thresholded accuracy, mean absolute error and mean absolute relative error as evaluation metrics. The definitions of these metrics are illustrated in Table 1.

## B. More Experiments

### Little-overlap Setting

Building on the work of NeuSurf, we validated the robustness of our method using the DTU little-overlap (Pixelnerf) setting. The little-overlap seltting is characterized by minimal overlap between the three images and larger pose intervals, making the reconstruction of geometry more challenging. As shown in Table 2, our approach achieved comparable experimental results in the little-overlap setting. This demonstrates the adaptability of our method to varying viewpoints in sparse settings, showcasing its universal sparse reconstruction capability.

### Dense View Reconstruction

To further substantiate the effectiveness of our method, we performed a comparative analysis against the current state-of-the-art neural implicit reconstruction and Gaussian reconstruction techniques with dense view input. Apart from some of the approaches we have discussed in the main text, additional works like NeRF (Mildenhall et al. 2020), NeuS2 (Wang et al. 2023b), SuGaR (Guédon and Lepetit 2023) and VCR-GauS (Chen et al. 2024b) are also compared. The results are illustrated in Table 3. The findings reveal that our

| Metric | Symbol | Definition |
| --- | --- | --- |
| Thresholded Accuracy | $< \delta$ (mm) | $\frac{1}{n}\sum_{i=1}^{n}[|\hat{y}_i - y_i| < \delta] \times 100\%$ |
| Mean Absolute Error | Abs. | $\frac{1}{n}\sum_{i=1}^{n}|\hat{y}_i - y_i|$ |
| Mean Absolute Relative Error | Rel. | $\frac{1}{n}\sum_{i=1}^{n}\left|\frac{\hat{y}_i - y_i}{y_i}\right| \times 100\%$ |

Table 1: Evaluation metrics for depth predition. $\hat{y}$ and $y$ denote predicted value and ground truth value, repectively.

| Scan ID | 24 | 37 | 40 | 55 | 63 | 65 | 69 | 83 | 97 | 105 | 106 | 110 | 114 | 118 | 122 | Mean |
| --- | --- | --- | --- | --- | --- | --- | --- | --- | --- | --- | --- | --- | --- | --- | --- | --- |
| COLMAP | 2.88 | 3.47 | 1.74 | 2.16 | 2.63 | 3.27 | 2.78 | 3.63 | 3.24 | 3.49 | 2.46 | 1.24 | 1.59 | 2.72 | 1.87 | 2.61 |
| SparseNeuS$_{ft}$ | 4.81 | 5.56 | 5.81 | 2.68 | 3.30 | 3.88 | 2.39 | 2.91 | 3.08 | 2.33 | 2.64 | 3.12 | 1.74 | 3.55 | 2.31 | 3.34 |
| VolRecon | 3.05 | 4.45 | 3.36 | 3.09 | 2.78 | 3.68 | 3.01 | 2.87 | 3.07 | 2.55 | 3.07 | 2.77 | 1.59 | 3.44 | 2.51 | 3.02 |
| UFORecon | 1.52 | 2.58 | 1.85 | 1.44 | 1.55 | 1.81 | 1.06 | 1.52 | 0.96 | 1.40 | 1.19 | 0.94 | 0.65 | 1.25 | 1.29 | 1.40 |
| NeuS | 4.11 | 5.40 | 5.10 | 3.47 | 2.68 | 2.01 | 4.52 | 8.59 | 5.09 | 9.42 | 2.20 | 4.84 | 0.49 | 2.04 | 4.20 | 4.28 |
| VolSDF | 4.07 | 4.87 | 3.75 | 2.61 | 5.37 | 4.97 | 6.88 | 3.33 | 5.57 | 2.34 | 3.15 | 5.07 | 1.20 | 5.28 | 5.41 | 4.26 |
| MonoSDF | 3.47 | 3.61 | 2.10 | 1.05 | 2.37 | 1.38 | 1.41 | 1.85 | 1.74 | 1.10 | 1.46 | 2.28 | 1.25 | 1.44 | 1.45 | 1.86 |
| NeuSurf | 1.35 | 3.25 | 2.50 | 0.80 | 1.21 | 2.35 | 0.77 | 1.19 | 1.20 | 1.05 | 1.05 | 1.21 | 0.41 | 0.80 | 1.08 | 1.35 |
| 2DGS | 3.25 | 3.64 | 3.52 | 1.42 | 2.04 | 2.52 | 1.99 | 2.69 | 2.55 | 1.79 | 2.92 | 4.50 | 0.73 | 2.38 | 1.79 | 2.52 |
| **Ours** | 1.32 | 2.85 | 2.71 | 0.80 | 1.44 | 2.08 | 1.11 | 1.19 | 1.33 | 0.76 | 1.49 | 0.85 | 0.47 | 1.05 | 1.06 | 1.37 |

Table 2: The quantitative comparison results of Chamfer Distance (CD↓) on DTU dataset (little-overlap setting). Cell colors denote best , second best and third best results.

| Scan ID | 24 | 37 | 40 | 55 | 63 | 65 | 69 | 83 | 97 | 105 | 106 | 110 | 114 | 118 | 122 | Mean |
| --- | --- | --- | --- | --- | --- | --- | --- | --- | --- | --- | --- | --- | --- | --- | --- | --- |
| NeRF | 1.90 | 1.60 | 1.85 | 0.58 | 2.28 | 1.27 | 1.47 | 1.67 | 2.05 | 1.07 | 0.88 | 2.53 | 1.06 | 1.15 | 0.96 | 1.49 |
| VolSDF | 1.14 | 1.26 | 0.81 | 0.49 | 1.25 | 0.70 | 0.72 | 1.29 | 1.18 | 0.70 | 0.66 | 1.08 | 0.42 | 0.61 | 0.55 | 0.86 |
| NeuS | 1.00 | 1.37 | 0.93 | 0.43 | 1.10 | 0.65 | 0.57 | 1.48 | 1.09 | 0.83 | 0.52 | 1.20 | 0.35 | 0.49 | 0.54 | 0.84 |
| NeuS2 | 0.56 | 0.76 | 0.49 | 0.37 | 0.92 | 0.71 | 0.76 | 1.22 | 1.08 | 0.63 | 0.59 | 0.89 | 0.40 | 0.48 | 0.55 | 0.70 |
| 3DGS | 2.14 | 1.53 | 2.08 | 1.68 | 3.49 | 2.21 | 1.43 | 2.07 | 2.22 | 1.75 | 1.79 | 2.55 | 1.53 | 1.52 | 1.50 | 1.96 |
| SuGaR | 1.47 | 1.33 | 1.13 | 0.61 | 2.25 | 1.71 | 1.15 | 1.63 | 1.62 | 1.07 | 0.79 | 2.45 | 0.98 | 0.88 | 0.79 | 1.33 |
| Gaussian Surfels | 0.66 | 0.93 | 0.54 | 0.41 | 1.06 | 1.14 | 0.85 | 1.29 | 1.53 | 0.79 | 0.82 | 1.58 | 0.45 | 0.66 | 0.53 | 0.88 |
| 2DGS | 0.48 | 0.91 | 0.39 | 0.39 | 1.01 | 0.83 | 0.81 | 1.36 | 1.27 | 0.76 | 0.70 | 1.40 | 0.40 | 0.76 | 0.52 | 0.80 |
| VCR-GauS | 0.55 | 0.91 | 0.40 | 0.43 | 0.97 | 0.95 | 0.84 | 1.39 | 1.30 | 0.90 | 0.76 | 0.92 | 0.44 | 0.75 | 0.54 | 0.80 |
| **Ours** | 0.40 | 0.76 | 0.30 | 0.31 | 0.85 | 0.80 | 0.56 | 1.36 | 0.92 | 0.62 | 0.53 | 0.66 | 0.32 | 0.38 | 0.40 | 0.61 |

Table 3: The quantitative comparison results of Chamfer Distance (CD↓) on DTU dataset (full-view reconstruction). Cell colors denote best , second best and third best results.

proposed method markedly improves reconstruction performance even when dealing with dense-view images.

**Direct Depth Fusion**

To explore the reconstruction quality of directly incorporating monocular depth into meshes, we conducted experiments on all 15 scenes of the DTU large-overlap setting. Notably, the monocular depth estimated by Marigold (Ke et al. 2024) is normalized and does not include real-world scale information. Therefore, applying a transformation to regularize the depth maps is necessary for proper fusion with ground truth poses. To maximize the performance, we calculate the scale and shift using ground truth depth maps through least squares fitting. The geometric evaluation re-

sults presented in Table 4 demonstrate that directly fusing monocular depth maps of multi-view images is not a practical way of reconstructing the surface.

**Novel View Synthesis**

We conduct few-shot novel view synthesis experiments on all the 15 tested DTU scenes. Following PixelNeRF (Yu et al. 2021), we use views 22, 25 and 28 of each scan as input and measure PSNR, SSIM (Wang et al. 2004) and LPIPS (Zhang et al. 2018) on 25 unseen images. We compare our approach with recent state-of-the-art methods including 3DGS, 2DGS, DNGaussian (Li et al. 2024) and FSGS (Zhu et al. 2023). The resolution of the evaluated images is 1/4 that of the original images. For a fair compari-

son, we use the same initial point clouds as ours for all the compared methods. For DNGaussian, we also report the results with randomly generated point clouds. The quantitative comparison results are reported in Table 5. Our approach achieves SOTA results, surpassing all the compared methods across the three evaluation metrics.

| Method | Mean CD↓ |
|---|---|
| Marigold (Ke et al. 2024) | 5.94 |
| Ours | **0.92** |

Table 4: Quantitative results on DTU dataset (large-overlap setting). The best result is marked as **bold**.

| Method | PSNR↑ | SSIM↑ | LPIPS↓ |
|---|---|---|---|
| 3DGS | 18.91 | 0.852 | 0.127 |
| 2DGS | 18.28 | 0.859 | 0.121 |
| DNGaussian (random) | 14.24 | 0.735 | 0.217 |
| DNGaussian | 17.85 | 0.792 | 0.170 |
| FSGS | 19.27 | 0.868 | 0.119 |
| **Ours** | 21.80 | 0.904 | 0.077 |

Table 5: Novel view synthesis results on DTU for 3 input views. Cell colors denote best, second best and third best results.

## C. Additional Ablation Study

In order to thoroughly evaluate the effectiveness and benefits of the modules we have proposed, we conducted a series of additional ablation studies. These studies involved systematically replacing our proposed modules with alternative approaches, as well as varying certain parameter settings of our method, to assess their individual contributions to the overall performance.

Directly employing monocular depth supervision faces a scale ambiguity between the real scene scale and the estimated depth. To mitigate this issue, FSGS (Zhu et al. 2023) introduces a relaxed relative loss $\mathcal{L}_p$ on the estimated and rendered depth maps, which measures the distribution difference between 2D depth maps and follows the below function:

$$\mathcal{L}_p = 1 - PCC(\widetilde{D}, \hat{D}). \quad (2)$$

Here, $\widetilde{D}$ and $\hat{D}$ represent the monocular depth map and the rendered depth map of a processing image, respectively. $PCC$ denotes the Pearson Correlation Coefficient, which is defined as

$$PCC(X,Y) = \frac{\text{Cov}(X,Y)}{\sqrt{\text{Var}(X)\text{Var}(Y)}}. \quad (3)$$

To investigate whether our purposed ranking constraint $\mathcal{L}_r$ is more effective for surface reconstruction, we replace $\mathcal{L}_r$ with $\mathcal{L}_p$ in our full model and measure the mean CD values of reconstructed meshes. As shown in Table 6, our ranking supervision $\mathcal{L}_r$ yields better results. This indicates that compared to the Pearson correlation optimization, our purposed ranking loss attains stronger feasibility on local geometry refinement. While $\mathcal{L}_p$ offers a general indication of global spatial distribution, it fails to address local errors and may introduce incorrect relative scale information.

| $\mathcal{L}_p$ | $\mathcal{L}_r$ | Mean CD↓ |
|---|---|---|
| √ |   | 1.24 |
|   | √ | **0.92** (Ours) |

Table 6: Comparison of surface reconstruction results using various types of monocular depth supervision on DTU dataset (large-overlap setting). The best result is marked as **bold**.

| $l=1$ | $l=2$ | Mean CD↓ |
|---|---|---|
| √ |   | 1.00 |
|   | √ | 1.01 |
| √ | √ | **0.92** (Ours) |

Table 7: Comparison of surface reconstruction results across different feature levels on DTU dataset (large-overlap setting). The best result is marked as **bold**.

We also explore the impact of varying multi-view feature levels on surface reconstruction. As outlined in the main text, we downsample the image resolution to $1/l$ of its original size before inputting it into the feature extraction network. We adopt a multi-level feature with $l = 1, 2$ in our full model. Compared to using either $l = 1$ or $l = 2$ alone, the introduction of multi-resolution image features enables more accurate correspondence of features between different viewpoints, thereby enhancing the reconstruction accuracy.

| Method | Mean CD↓ |
|---|---|
| Model A | 2.10 |
| Model B | 2.33 |
| Ours w/o $\mathcal{L}_r, \mathcal{L}_s$ | 1.93 |
| Ours w/o $v$ | 1.62 |
| Ours w/o patches | 4.26 |
| Ours | **1.37** |

Table 8: Comparison of surface reconstruction results on DTU dataset (little-overlap setting). The best result is marked as **bold**.

To investigate if absolute depth or multi-view color supervision outperforms our proposed method, we conduct ablation study on the two losses. Reconstruction CD results on DTU (little-overlap setting) are shown in the Table 8. *Model A* refers to the model obtained by replacing the depth ranking loss with the scale-shifted MSE loss (from MonoSDF

| Scan ID | 24 | 37 | 40 | 55 | 63 | 65 | 69 | 83 | 97 | 105 | 106 | 110 | 114 | 118 | 122 | Mean |
|---|---|---|---|---|---|---|---|---|---|---|---|---|---|---|---|---|
| Geo-NeuS | 3.55 | 13.99 | 3.03 | 0.77 | 1.56 | 5.01 | 0.74 | 1.16 | 1.00 | **0.68** | 0.84 | 0.52 | **0.41** | 0.78 | 0.92 | 2.33 |
| DN-Splatter | 8.74 | 9.51 | 8.62 | 8.51 | 8.40 | 7.34 | 7.42 | 8.17 | 8.93 | 8.20 | 9.90 | 8.42 | 8.08 | 7.95 | 7.02 | 8.35 |
| PGSR | 3.17 | 5.02 | 2.72 | 2.59 | 8.74 | 3.85 | 2.61 | 3.13 | 7.27 | 3.04 | 2.47 | 4.34 | 2.60 | 2.99 | 2.88 | 3.83 |
| SparseCraft | 1.17 | **1.74** | 1.80 | **0.70** | **1.19** | 1.53 | 0.83 | **1.05** | 1.42 | 0.78 | 0.80 | 0.56 | 0.44 | **0.77** | **0.84** | 1.04 |
| **Ours** | **0.67** | 1.94 | **1.17** | 0.77 | 1.28 | **1.23** | **0.63** | **1.05** | **0.98** | 0.69 | **0.75** | **0.48** | **0.41** | 0.78 | 0.90 | **0.92** |

Table 9: The quantitative comparison results of Chamfer Distance (CD↓) on DTU dataset (large-overlap setting). The best results are marked as **bold**.

(Yu et al. 2022)) in our full model. *Model B* refers to the model obtained by replacing the feature loss with multi-view multi-resolution color loss in our full model. The results confirm that it is challenging to obtain precise scale and shift with sparse views due to the absence of multi-view constraints. And the multi-view multi-resolution color loss is affected by reflective non-Lambertian surfaces.

We also conduct ablation study on the visibility item $v$, the feature loss $\mathcal{L}_f$ and the patches used in depth ranking loss $\mathcal{L}_r$, as shown in Table 8. Incorporating $\mathcal{L}_f$ results in a 0.54 improvement in CD (2.47 to 1.93) while $\mathcal{L}_r$ and $\mathcal{L}_s$ together provide an 0.56 improvement (1.93 to 1.37). While the use of patches for depth ranking loss ensures that pixels involved in the loss calculation are not too far, thus reducing the impact of inaccurate values in monocular depth estimation, it introduces a trivial problem. Specifically, a pixel on the edge of a patch cannot use its immediate neighbor in an adjacent patch for ranking loss calculation, leading to noise at the edges of the patches. To address this issue, we use the strategy of overlapping patches in our implementation.

## D. More Comparisons

### Comparison with Surface Reconstruction Methods

Besides the methods compared in the main text, there are some other noteworthy relevant works in the existing publications and concurrent submissions: Geo-NeuS (Fu et al. 2022), DN-Splatter (Turkulainen et al. 2024), PGSR (Chen et al. 2024a) and SparseCraft (Younes, Ouasfi, and Boukhayma 2024).

- **Geo-NeuS in NeurIPS 2022** is the state-of-the-art neural-rendering method in dense view reconstruction on DTU dataset. It directly locates the zero-level set of SDF networks and explicitly perform multi-view geometry optimization by leveraging the sparse geometry from SfM and photometric consistency in multi-view stereo.
- **DN-Splatter in arXiv 2024** extends 3D Gaussian splatting with depth and normal cues for efficient mesh extraction. It regularizes the optimization procedure with depth information and uses the geometry of the 3D Gaussians supervised by normal cues to achieve better alignment with the true scene geometry.
- **PGSR in arXiv 2024** is a state-of-the-art method for dense view surface reconstruction based on the Gaussian Splatting pipeline. It leverages geometric regularization and multi-view photometric consistency to optimize the reconstruction results.
- **SparseCraft in arXiv 2024** leverages the neural rendering framework (similar to NeuS), incorporating stereopsis cues to improve the optimization of geometric and color fields. Additionally, CUDA optimization is used to accelerate the optimization process.

We compared our method with these methods in terms of sparse reconstruction quality under the DTU large-overlap setting, as shown in Table 9. Our reconstruction results significantly outperform those of contemporary methods. Similar to our approach, PGSR employs multi-view priors and leverages consistency across different viewpoints for optimization. However, as discussed in our main text, PGSR relies solely on multi-view photometric consistency, which only works well with dense view inputs. Unfortunately, its performance degrades sharply under sparse input conditions due to the influence of lighting variations. SparseCraft continues to utilize the neural radiance fields method, resulting in a reconstruction quality that is marginally inferior to our approach. Besides, in contrast to our Gaussian Splatting-based technique, the NeRF-based method encounters significant difficulties in further optimizing both rendering speed and quality, thereby precluding the possibility of real-time rendering.

### Comparison with Novel View Synthesis Methods

There exist a few papers in the literature that focus on regularizations for novel view synthesis with sparse input views, some of which can be trivially incorporated for surface reconstruction: ViP-NeRF (Somraj and Soundararajan 2023), SparseNeRF (Wang et al. 2023a), DSNeRF (Deng et al. 2022), DDP-NeRF (Roessle et al. 2022) and FSGS (Zhu et al. 2023). We implemented the core ideas of these related works on 2DGS baseline. Reconstruction CD results on DTU (large-overlap setting) are shown in Table 10. We hope this sets a proper baseline for future work.

The experimental results indicate that our method outperforms others by incorporating ideas borrowed from novel view synthesis works. ViP-NeRF improves training with sparse input views by using a visibility prior through the use of plane sweep volumes. However, it exhibits significant depth estimation errors in regions where only a single view is visible, which is more problematic in geometric reconstruction tasks. DSNeRF and DDP-NeRF directly optimize the rendered depth using the absolute values from SfM depth/completed depth. These methods require a high level of completeness and accuracy in depth prediction, which currently limits their applicability in sparse-view reconstruc-

| Method | Mean CD↓ |
|---|---|
| ViP-NeRF-2DGS | 2.55 |
| SparseNeRF-2DGS | 3.14 |
| DS-NeRF-2DGS | 2.39 |
| DDP-NeRF-2DGS | 1.91 |
| FSGS-2DGS | 1.80 |
| Ours | **0.92** |

Table 10: Comparison of surface reconstruction results on DTU dataset (large-overlap setting). The best result is marked as **bold**.

tion. The depth loss in SparseNeRF, similar to our approach, employs a ranking relationship to avoid introducing additional errors. However, this can result in overly smooth object surfaces lacking reliable details. FSGS utilizes the Pearson correlation coefficient to optimize rendering depth under sparse views. We conducted similar procedures in our ablation experiments, which resulted in a deterioration of local geometry.

## E. Applications in Other Fields

Our method is a general sparse-view object reconstruction approach that incorporate intra-view depth and multi-view feature consistency to 2D Gaussian Splatting pipeline for fast and accurate surface reconstruction. In certain more specialized fields, there is also a demand for sparse-view reconstruction. In the process of reconstructing the human head, it is typically feasible to acquire only a limited number of images through low-cost equipment. H3DS (Ramon et al. 2021), deformable model-driven approaches (Xu et al. 2023) and Implicit Neural Deformation methods (Li et al. 2022) integrate neural implicit fields with prior knowledge models of the head to achieve high-fidelity head reconstruction. However, the optimization of neural implicit fields still requires a significant amount of time. In the future, extending our method to sparse-view reconstruction tasks for the head or other objects with prior shape information may be an effective solution to enhance reconstruction efficiency while ensuring quality.

## F. Potential Negative Social Impacts

Our approach utilizes the Gaussian Splatting pipeline to rapidly reconstruct general objects from sparse views. However, the resulting reconstructions may pose potential copyright issues. Rapidly reconstructing objects without proper authorization can complicate copyright management and enforcement. Our reconstruction process necessitates the use of high-performance GPUs to perform the computations, which may lead to considerable energy consumption and potential environmental impact.

## G. More Qualitative Results

In our main text, we show several representative scenes for visual comparison on DTU dataset. In this supplementary, we present the qualitative reconstruction results on all 15 tested scenes from the DTU dataset under different overlap settings in Figure 1 and Figure 2. Figure 1 highlights the results in the large-overlap setting, while Figure 2 illustrates the results in the little-overlap setting. These visualizations provide a thorough examination of the performance across different overlap conditions.

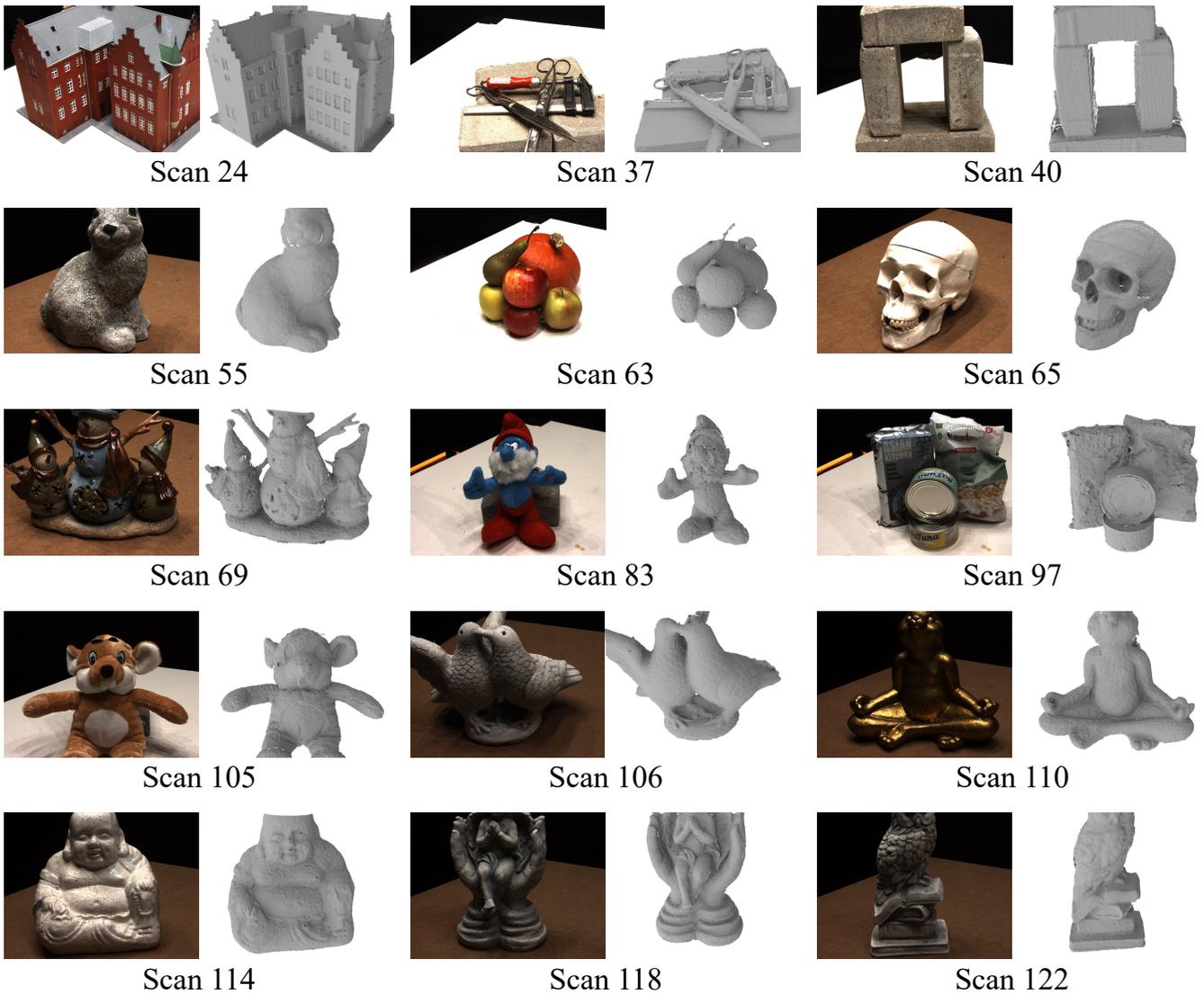

Figure 1: Visualized results of 3-view surface reconstruction on DTU dataset (large-overlap setting).

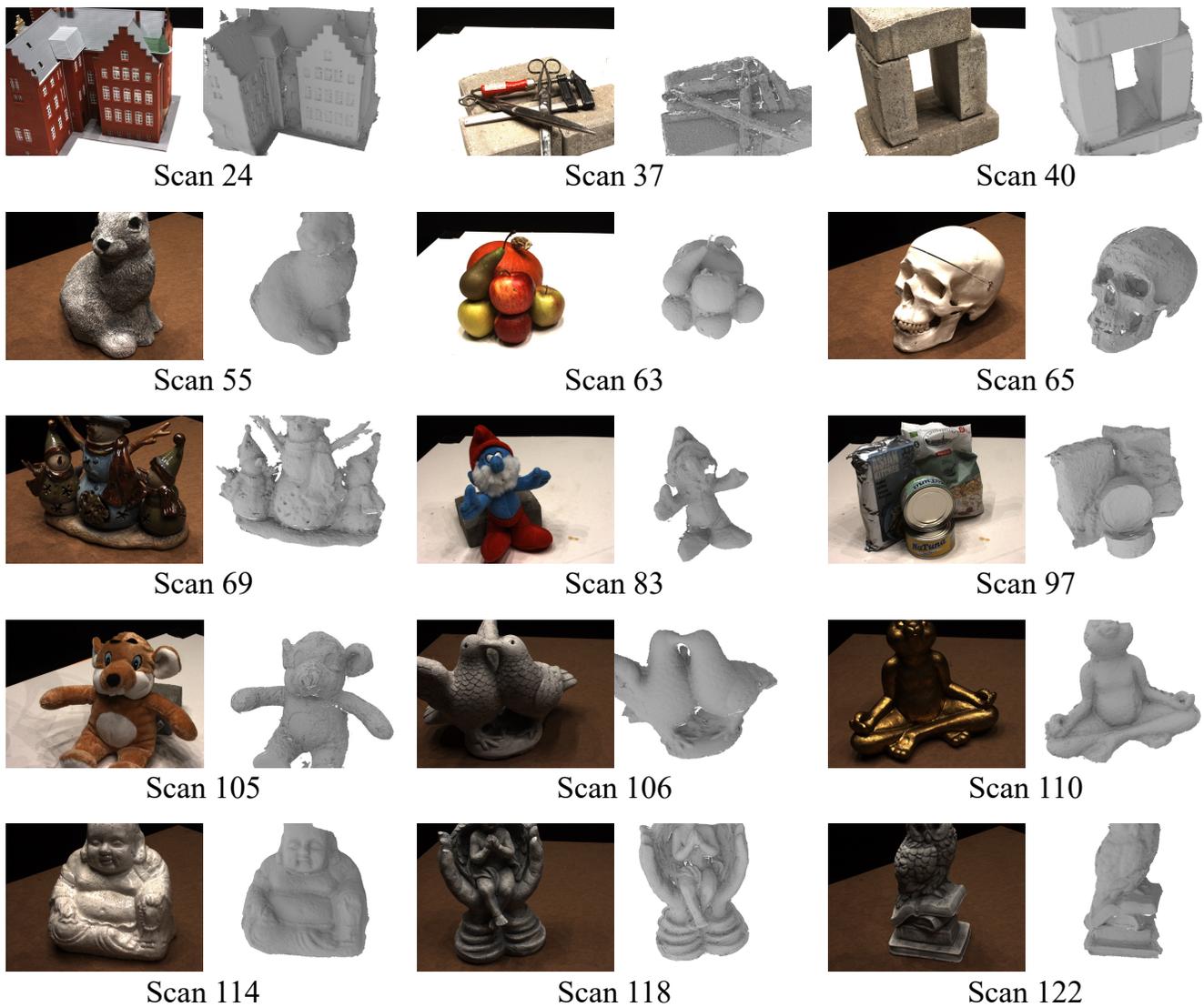

Figure 2: Visualized results of 3-view surface reconstruction on DTU dataset (little-overlap setting).



\end{document}